\documentclass{article}

\PassOptionsToPackage{numbers, compress}{natbib}

\usepackage[top=1in, bottom=1in, left=1in, right=1in]{geometry}

\usepackage[utf8]{inputenc} 
\usepackage[T1]{fontenc}    
\usepackage{hyperref}       
\usepackage{url}            
\usepackage{booktabs}       
\usepackage{amsfonts}       
\usepackage{bm}
\usepackage{nicefrac}       
\usepackage{microtype}      
\usepackage{enumitem}
\usepackage{epstopdf}
\usepackage[norelsize,boxed,linesnumbered,vlined,algo2e]{algorithm2e} 
\usepackage{algorithm}
\usepackage{algorithmic}		
\usepackage[table,xcdraw]{xcolor}					

\newtheorem{definition}{Definition}

\usepackage{amsmath}
\usepackage{amsfonts}
\usepackage{amssymb}
\usepackage{wrapfig}
\usepackage{subcaption}
\usepackage{multirow}
 \usepackage{mathtools} 

\usepackage{verbatim}

\usepackage{anyfontsize}

\usepackage{color}
\usepackage{tikz}
\usetikzlibrary{arrows,shapes,snakes,automata,backgrounds,fit,petri}

\newcommand{\RN}[1]{%
	\textup{\lowercase\expandafter{\it \romannumeral#1}}%
}

\usepackage{algorithm}
\usepackage{algorithmic}

\newcommand{\ie}[0]{\emph{i.e., }}

\newcommand{\eg}[0]{\emph{e.g., }}

\newcommand{\beq}{\vspace{0mm}\begin{equation}}
\newcommand{\eeq}{\vspace{0mm}\end{equation}}
\newcommand{\beqs}{\vspace{0mm}\begin{eqnarray}}
\newcommand{\eeqs}{\vspace{0mm}\end{eqnarray}}
\newcommand{\barr}{\begin{array}}
\newcommand{\earr}{\end{array}}

\newcommand{\av}[0]{{\boldsymbol{a}}}
\newcommand{\bv}[0]{{\boldsymbol{b}}}
\newcommand{\cv}[0]{{\boldsymbol{c}}}

\newcommand{\fv}[0]{{\boldsymbol{f}}}

\newcommand{\tv}[0]{{\boldsymbol{t}}}
\newcommand{\uv}{\boldsymbol{u}}

\newcommand{\yv}{\boldsymbol{y}}
\newcommand{\zv}{\boldsymbol{z}}

\newcommand{\phiv}{\boldsymbol{\phi}}

\newcommand{\R}{\mathbb{R}}

\newcommand{\N}{\mathcal{N}}

\ifx\assumption\undefined
\newtheorem{assumption}{Assumption}
\fi

\ifx\definition\undefined
\newtheorem{definition}{Definition}
\fi

\ifx\remark\undefined
\newtheorem{remark}{Remark}
\fi

\usepackage{xr}
\usepackage{graphicx}
\graphicspath{{./Pictures/}}

\title{PiPs: a Kernel-based Optimization Scheme for Analyzing Non-Stationary 1D Signals
}

\author{
Jieren Xu$^1$, Yitong Li$^2$, Haizhao Yang$^3$, \\
David Dunson$^4$,  Ingrid Daubechies$^4$ \\
 ${}^1$ Google LLC, ${}^2$ Apple Inc, ${}^3$Duke University, ${}^4$University of Maryland College Park\\
  \texttt{${}^3$jierenxu@google.com, ${}^2$yitongli@apple.com, ${}^3$ingrid,dunson@duke.com}\\ \texttt{ ${}^4$hzyang@umd.edu } \\
}

\begin{document}

\maketitle

\begin{abstract}
This paper proposes a novel kernel-based optimization scheme to handle tasks in the analysis, \eg signal spectral estimation and single-channel source separation of 1D non-stationary oscillatory data.
The key insight of our optimization scheme for reconstructing the time-frequency information is that when a nonparametric regression is applied on some input values, the output regressed points would lie near the oscillatory pattern of the oscillatory 1D signal only if these input values are a good approximation of the ground-truth phase function.
In this work, \textit{Gaussian Process (GP)} is chosen to conduct this nonparametric regression:
the oscillatory pattern is encoded as the \textit{Pattern-inducing Points (PiPs)} which act as the training data points in the GP regression;  while the targeted phase function is fed in to compute the correlation kernels, acting as the testing input.
Better approximated phase function generates more precise kernels, thus resulting in smaller optimization loss error when comparing the kernel-based regression output with the original signals.
To the best of our knowledge, this is the first algorithm that can satisfactorily handle fully non-stationary oscillatory data, close and crossover frequencies, and general oscillatory patterns. 
Even in the example of a signal {produced by slow variation in the parameters of a trigonometric expansion}, we show that PiPs admits competitive or better performance in terms of accuracy and robustness than existing state-of-the-art algorithms.  
\end{abstract}

\section{Introduction}
\label{sec:intro}
This paper is concerned with 1D single-channel source separation and estimation for oscillatory signals. Suppose a signal $f(t)$ is defined on a time domain $[0,T]$ with $K$ intrinsic components and non-constant frequencies:
\beq
\label{eqn:signal}
f(t)=\sum_{k=1}^K f_k(t) =\sum_{k=1}^K a_k(t) s_k( \phi_k(t)),
\eeq
where $a_k(t)$ and $\phi_k(t)$ are smooth, slowly varying functions representing the latent \textit{amplitude} and \textit{phase functions} of the $k$th component, $f_k(t)$, for $k=1,\ldots,K$. The derivative of {phase function} $\phi_k(t)$ is called the \textit{frequency function}, denoted as $\omega_k(t)$ and is also assumed to be smooth. 
$s_k(t)$ is a periodic \textit{shape} (or \textit{pattern}) function for the $k$th component, describing a potentially complicated evolution pattern of the signal. 
We assume $s_k(t)$ to be bounded, continuous, to have periodicity 1 and to satisfy $\int_0^1 s_k(t)dt=0$, with unit $L_2$-norm on $[0,1]$. 
The variation of $a_k(t)$ and $\omega_k(t)$ are assumed to be sufficiently small and the magnitude of $\omega_k(t)$ is assumed to be large enough such that the pattern is well defined. 

\begin{figure}
\centering
  \includegraphics[height=3.5in]{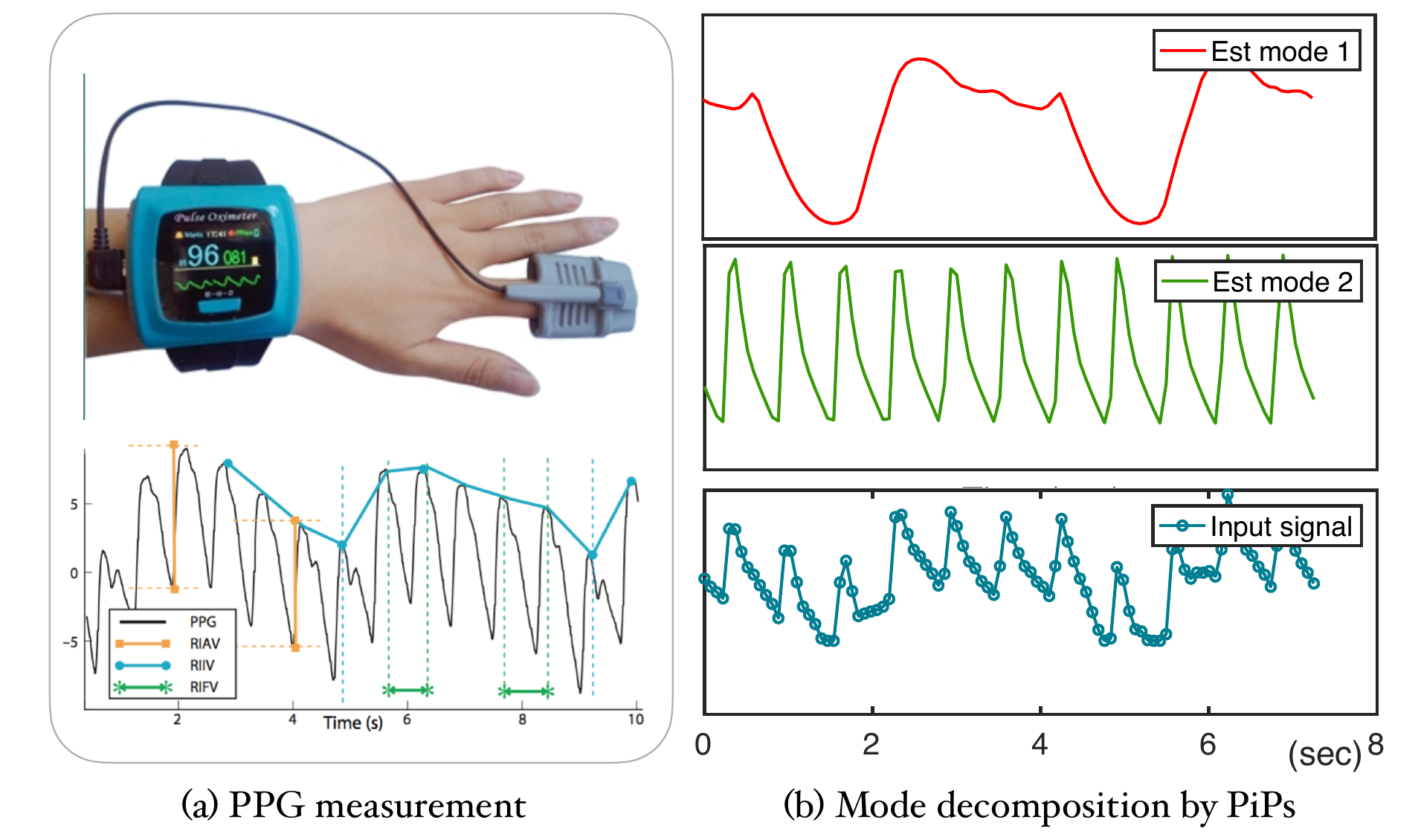}  
\caption{\footnotesize{ (a) Measurement and several explicitly handcrafted properties of Photoplethysmogram (PPG) signal {as a motivating example of non-stationary 1D signals}. (b) Reconstructed components (top two figures  for Fig 1(b)) of the PPG signal (bottom figure of Fig 1(b)). These two components were reconstructed from only a small portion (100 points) of the samples of the original PPG raw data as visualized in the bottom figure.}}
\label{fig6}
\end{figure}

One toy example of Model~\eqref{eqn:signal} are trigonometric functions. 
A more complicated example in (Fig~\ref{fig6}(b:bottom)) is, \eg a real Photoplethysmogram (PPG) signal {in Figure \ref{fig6}(a)}; the PPG signal describes the human cardiac and respiratory cycles with $K=2$ intrinsic components: the first component (Fig~\ref{fig6}(b:middle)) represents the beating of the heart and the second represents the cyclic respiratory behavior of the lungs (Fig~\ref{fig6}(b:top)). 
Model~\eqref{eqn:signal} includes a large family of approximately periodic signals in real applications \cite{Eng1,Eng2,SSCT,Canvas,HauBio2,Pinheiro2012175,Crystal,Eng2,ME,Canvas,Canvas2,GeoReview,SSCT,MUSIC,SuperResolution:Candes,gruber1997statistical,burg1972relationship,roy1989esprit}.

Solving Equation~\eqref{eqn:signal} {(i.e., identifying amplitude, phase, and shape functions from $f(t)$ in Equation \eqref{eqn:signal})} is a general task that involves several sub-problems: (i) spectral estimation
when $\omega_k(t)$ is linear; (ii) adaptive time-frequency analysis\cite{auger1995improving,Daubechies1996,YANG2017} that aims to retrieve time-variant information $a_k(t)$, $\phi_k(t)$, $\omega_k(t)$; 
(iii) mode decomposition~\cite{Huang1998,Wu2009,Wu2009EEMD} that targets the extraction of $f_k(t)$; 
(iv) pattern recognition~\cite{zhu2013locally} to reconstruct $s_k(t)$, etc.
Generally, $f(t)$ and $K$ are fed as input information for the above-mentioned approaches.

Despite many successful algorithms for solving these sub-problems, to the best of our knowledge, no algorithm in the literature satisfactorily fulfill the ultimate goal of estimating $a_k(t)$, $\phi_k(t)$ (or $\omega_k$(t)), and $s_k(t)$ when $f(t)$ is fully non-stationary with close and crossover frequencies, and general patterns.
Moreover, many existing algorithms require a high sampling rate, which is not always practical (\textit{e.g.}) for oscillatory data collected by mobile devices, such as portable health monitors (see Figure~\ref{fig6}(a)), due to the limit of battery capacity.


This paper proposes a framework that can estimate $a_k(t)$, $\phi_k(t)$ (or $\omega_k(t)$), and $s_k(t)$ simultaneously from relatively few samples of $f(t)$. The algorithm requires a prior input of (1) the number of intrinsic components $K$; (2) a rough estimate of the frequency and pattern functions. 
The estimate in (2) can be quite rough: for instance, for the PPG signal in Figure~\ref{fig6}, we initialize the patterns for both components as a sine function, with frequencies of $15$ periods/min and $95$ beats/min, which is out of common-knowledge rule-of-thumb approximations; the output gives the respective shapes and frequencies for the heart and lung components with the desired accuracy (Figure~\ref{fig6}(b)).

Our framework applies a two-stage iteration scheme until convergence: one stage to update phase functions (and amplitudes) and the other stage to update oscillatory patterns. The phase updating stage is the core part of the algorithm. 
There are four components for nonparametric regression: the input and output of the training points and the input and output of the testing points.
Our key intuition is that when a nonparametric regression is applied on some input values, the output regressed points would lie near the oscillatory pattern only if these input values are a good approximation of the ground-truth phase function.
The nonparametric regression here is implemented by the \textit{Gaussian Process (GP)}, since the GP-regression-based implementation shows more robustness compared to several other standard nonparametric regression approaches~\cite{gyorfi2002distribution}.

 In this stage, first, we encode the prior knowledge of the patterns using the \textit{Pattern-inducing Points (PiPs)}. Then we formulate a GP-regression-based optimization problem to retrieve the phase functions by treating the PiPs as training (input-output) points, while the phase functions as the latent testing input and the original signal samples as the testing output respectively.
As the input of a GP, the targeted phase function is fed into the correlation kernels to compute the output values of the regression.
Since better-approximated phase function generates more precise kernels, thus resulting in a smaller difference between the point-wise kernel-based regression output and the original signals, we design the optimization loss as the $L_2$ distance between regression output and the noisy measurement of the 1D signal.
By optimizing this innovative kernel-based loss function, the latent input phase function is retrieved.
In this sense, this stage can also be viewed as a latent GP regression problem that aims to recover the latent input of a GP given the output values.

To enhance the performance of the kernel-based optimization, we transform the nonparametric setting to a semi-parametric setup by deploying a divide-and-conquer strategy.
We separate the long signals into multiple localized signal chips. These chips are supported on continuous time intervals which can have intersections, as long as the whole time span of the original signal is fully covered.
In the phase-updating stage, since each signal chip is time-localized and the variation of the time-instantaneous information is assumed to be sufficiently small, we propose to use low-order polynomials to model the phase and amplitude functions for each of these local chips.
By transforming the original nonparametric model to the current semi-parametric one, we can guarantee the local monotonicity and the smoothness of the time-instantaneous information, thus largely enhancing the robustness of the optimization process. 
Then we summarize the time-instantaneous information for all chips and feed it into the pattern-updating stage to update the oscillatory patterns.

In the pattern-updating stage, state-of-the-art 1D pattern recovery algorithms, \eg Recursive Diffeomorphism-Based Regression for Shape
Functions (RDBR)~\cite{xu2018recursive}, are applied to update the oscillatory patterns given the renewed time-frequency information and the noisy measurement of the 1D oscillatory signal.
This stage allows us to handle oscillatory signals with a broad class of oscillatory patterns as long as a rough initialization is provided, compared to the traditional methods that are mostly limited to trigonometric oscillatory patterns.

The first and second stages of the algorithm are introduced in Sections~\ref{sec:MI} and~\ref{sec:Shape}. Section~\ref{sec:pgpr} summarizes PiPs. As we shall see in the numerical examples in Section~\ref{sec:num}, PiPs works for a wide range of signals in Model~\eqref{eqn:signal} while existing methods fail for certain or all aspects. Moreover, for simple signals that can be handled by super-resolution analysis, PiPs achieves better results compared to several state-of-the-art methods with only reasonable initial values. 


%


\section{Estimation of Phase Functions} \label{sec:MI}
This section explains how to update the phase and amplitude functions in Model~\eqref{eqn:signal} when exact or approximated knowledge of the pattern $s_k(t)$ is provided. 

To fix our thoughts, we start by introducing the concept of PiPs; these are on-grid auxiliary points that play a role similar to training points in other learning processes. {To be specific, we introduce a formal definition as follows.}
\begin{definition}\textbf{(PiPs)}
Let $s(t)$ be a periodic function that satisfies the assumptions in Section~\ref{sec:intro}. We say that the points $(P_i)_{i=0}^{l}$, with coordinates $(z_i, u_i)\in \R^2$ for all $i\in\{0,\ldots,l\}$ are \textit{Pattern-inducing Points (or PiPs)} for $s(t)$,
with tolerance $h$ on the interval $[b_0,b]\subset \R$, where $1\leq b_0 - b < \infty $, if 
\begin{enumerate}[label=\alph*)]
    \item  $z_i = b_0 + \frac{b - b_0}{l} i $
    \item the continuous affine functions $\tilde{s}: [b_0, b] \to \R $ with breakpoints at the $z_i$, and such that $\tilde{s}(z_i) = u_i$, $i = 0, \ldots, l$, satisfies $\|(\tilde{s} - s)|_{[b_0,b]}\|_{\infty} \leq h$.
\end{enumerate}
\end{definition}

\begin{figure}
\centering
  \includegraphics[height=1in]{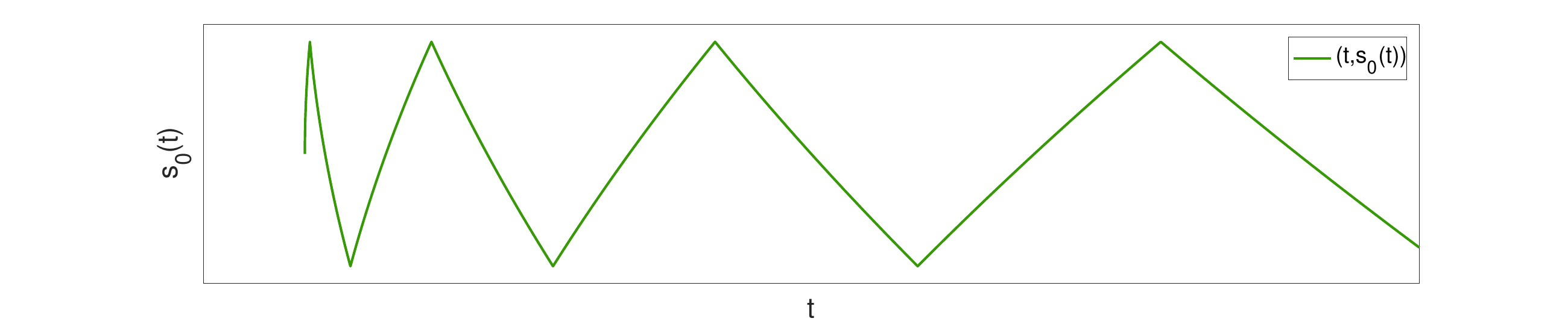} \\  
  (a) Cartoon for one component $s_k(\phi_k(t))$ in Model~\eqref{eqn:signal}. \\
    \includegraphics[height=1in]{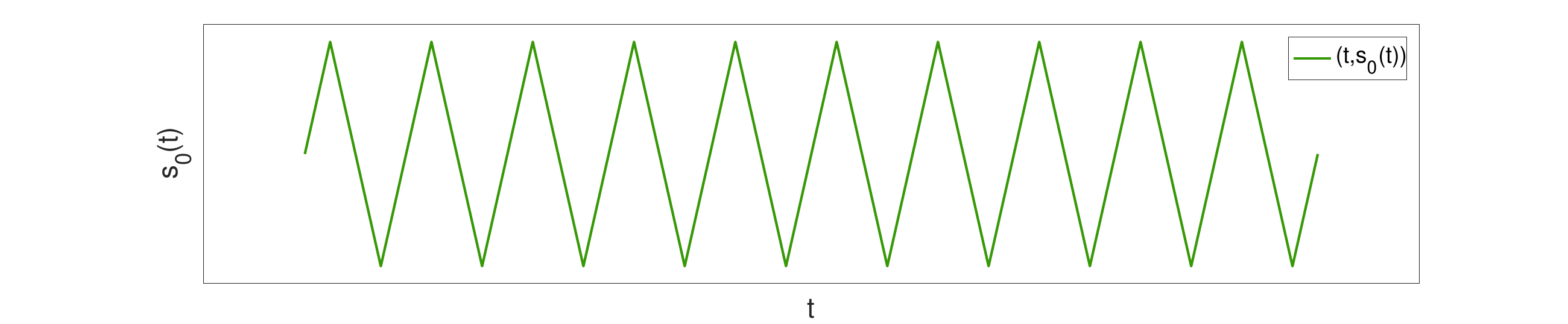} \\
    (b) Periodic pattern  $s_k (t)$. \\
    \includegraphics[height=1.4in]{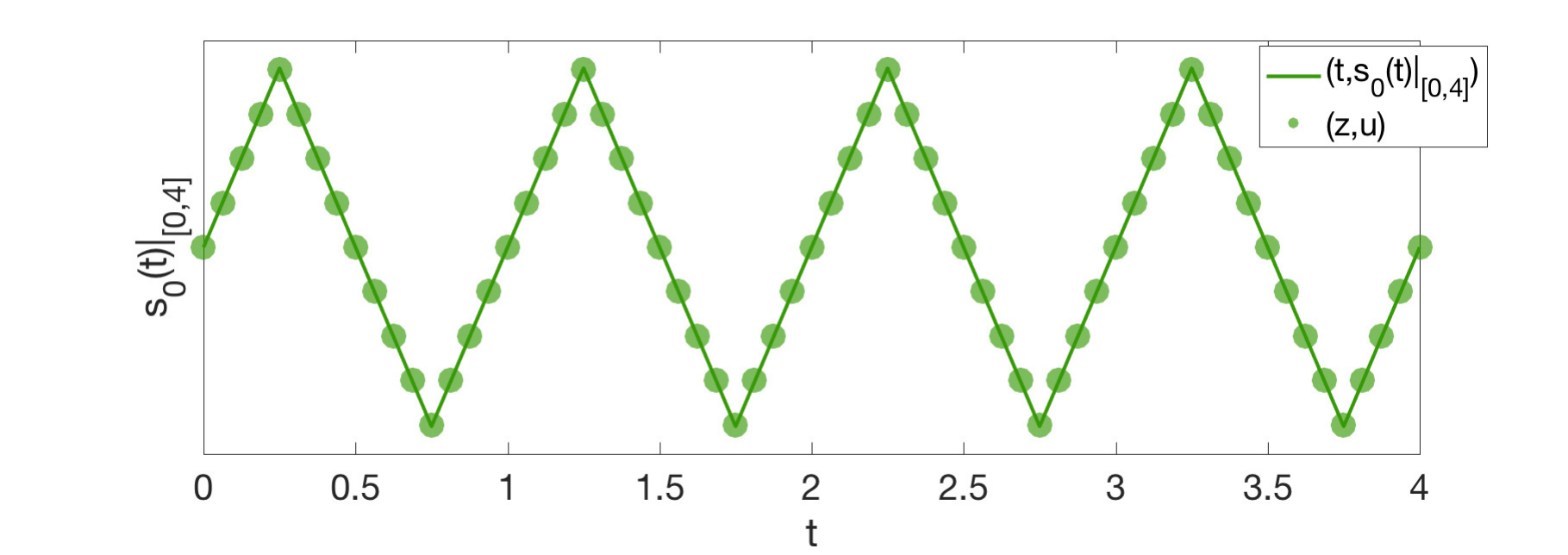}  \\
(c) An example of PiPs (green dots) for $s_k (t)$.
\caption{\footnotesize{ The original oscillatory pattern and two non-oscillatory patterns with no approximation error.}}
\label{nop}
\end{figure}

Figure~\ref{nop}(a) shows a cartoon for one component $s_k(\phi_k(t))$ in Model~\eqref{eqn:signal}. Figure~\ref{nop}(b) shows one corresponding periodic pattern  $s_k (t)$, which is also the unwarping result of $s_k(\phi_k(t))$ w.r.t. $\phi_k(t)$.
Figure~\ref{nop}(c) is an example of PiPs (green dots) for $s_k (t)$ on $[0,4]$  with $l = 16$.

Note that for standard signals with trigonometric patterns, the PiPs is considered to be already known. In practise, we set $b_0 = 0$  without loss of generality. The pattern resolution $l$ should be large enough to describe the details of each pattern. Moreover, the left-hand bound $b$ should be larger than the upper bound of $\phi_k(t)$ to guarantee the optimization performance.

Next by fixing the PiPs as the training points, a nonparametric-regression-based optimization algorithm is designed to retrieve the latent phase and amplitude functions by maximizing the posterior distribution of the observed samples.

\subsection{GP Regression}
\label{sec:reg}
{
Suppose $\yv\in\R^N$ are observations\footnote{We will use bold font for vectors and $(\cdot)_i$ for the $i$th element of the respective vector.} of
 \begin{align} 
 y(t)= \sum_k^K y_k(t),\ \text{ where  } \ \ y_k(t) = f_k(t) + ns_k(t),  \nonumber
 \end{align}
 sampled at time points $\tv=[t_1,\dots,t_N]\in\R^N$. Here we consider $ns_k(t)$ as GP with mean 0 and fixed point-wise variance $\sigma_k^2$. Thus $y(t)$, being the sum of K independent GP's, can also be modeled as a GP and tackled with respective tools.
 In the rest of this section, we illustrate the key idea of this work, \textit{i.e.}, using PiPs ($y(\phi)$) to formulate this nonparametric regression problem, rather than modeling $y(t)$ directly. }

\begin{figure}
\centering
  \includegraphics[height=1.5in]{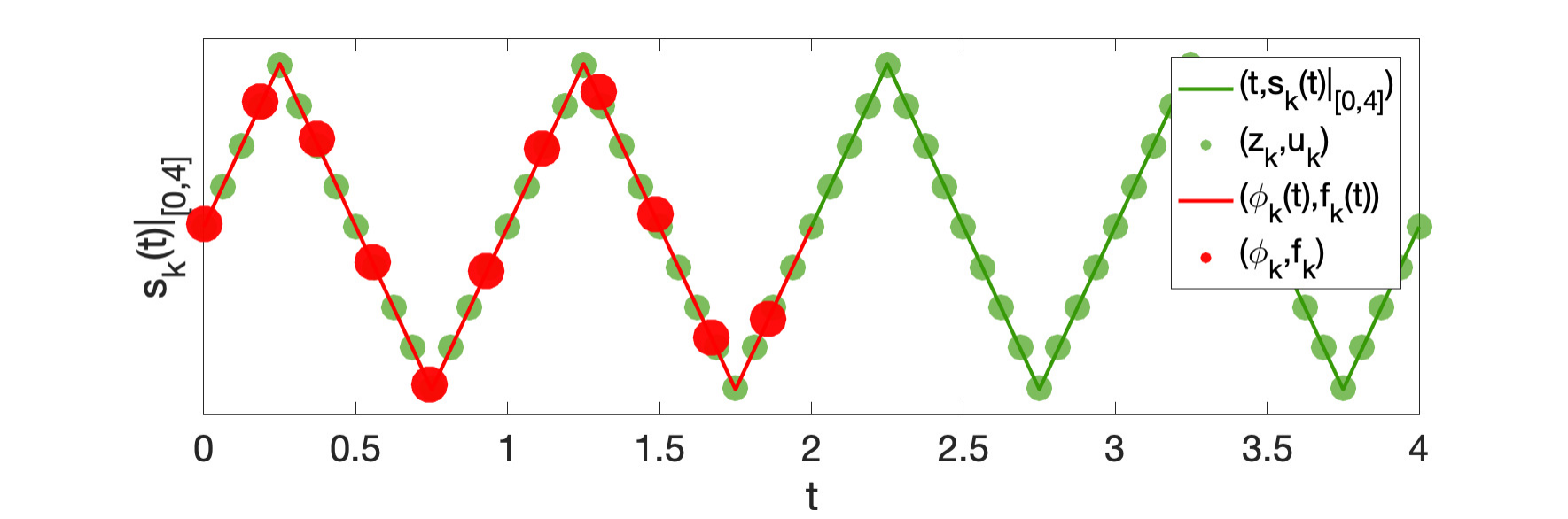}    \includegraphics[height=1.5in]{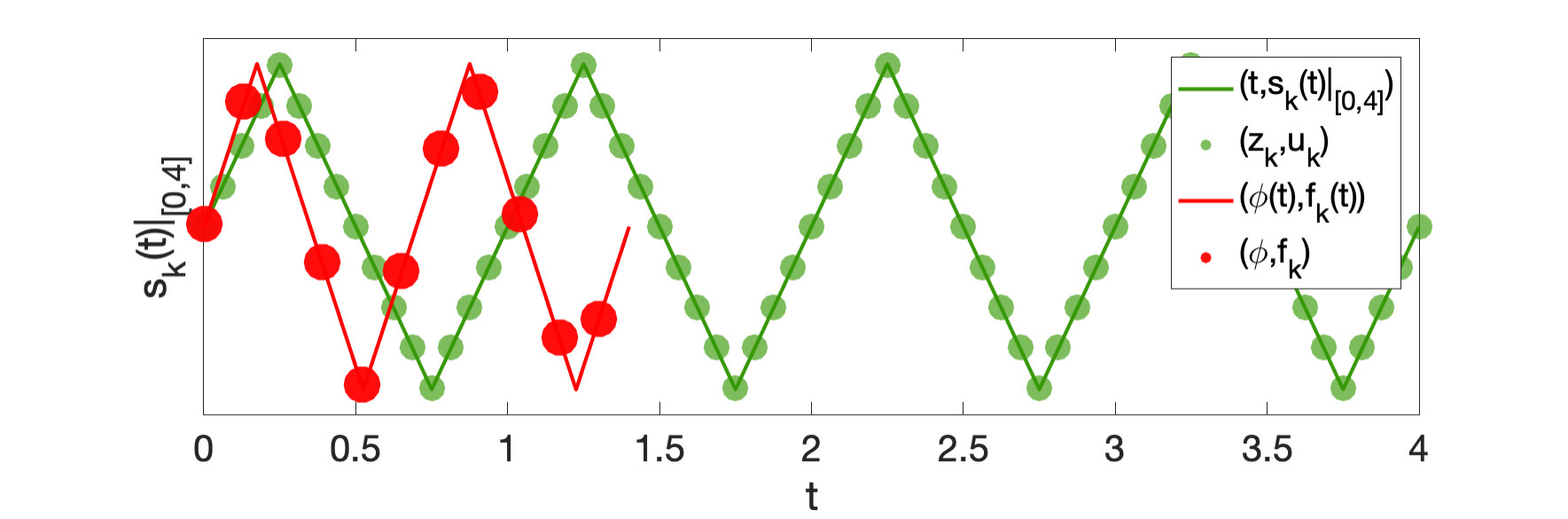}  \\
(a) $(\phiv_k,\fv_k)$ lies on $s_k(t)$.  \hspace{2cm} 
(b) $(\phiv,\fv_k)$ lies off  $s_k(t)$.
\caption{\footnotesize{ Phase-signal pairs (red points) that lies on or off the $s_k(t)$ (green curve). (a) When the phase is set to the ground truth value $\phiv_k$, the signal points $\fv_k$ lie on $s_k(t)$. Hence non-parametric regression  can be used to estimate the signal value (red points) from the PiPs  $(\zv_k,\uv_k)$ (green points). 
(b) When the phase deviates from ground truth, non-parametric regression can't be used to estimate signal value (red points) given the PiPs (green points).
}}
\label{reg}
\end{figure}

For each mode $f_k(t) = a_k(t) s_k(\phi_k(t))$, denote $\av_k = a_k(\tv)$, $\phiv_k = \phi_k(\tv)$ and $\fv_k = \av_k \odot s_k(\tv)$ as the respective discretization\footnote{$\odot$ is entry-wise product between vectors or matrix.}
of $a_k(t)$, $\phi_k(t)$ and $f_k(t)$ at time samples $\tv$. 
It's easy to see that if we set $a_k(t)$ to a constant say $a_k(t) = 1$, the re-arranged mode points  $(\phiv_k,\fv_k ) = (\phiv_k,s_k(\phiv_k))$ should  lie exactly on the oscillatory pattern $s_k(t)$. On the other hand, since $\phi_k(t)$ is a strictly monotonic function, for any $\phi(\tv)$ that deviates from $\phi_k(\tv)$ module $1$, \ie $\phi(\tv)\neq \phi_k(\tv) \mod 1$, $(\phiv,\fv_k)$ should deviate from pattern $s_k(t)$. This is illustrated in Figure~\ref{reg} (a)(b) respectively, where $s_k(t)$ is has a triangular pattern and $\phi_k(t) = 2t$, $t\in [0,1]$.

As we know, there are four essential components for nonparametric regression: the input and output of the training points, and the input and output of the testing points.
Based on the aforementioned observation, if we set $(\zv_k,\uv_k)$'s as PiPs for $f_k(t)$, $[\min \phiv_k,\max \phiv_k]\in [b_0,b]$, and treat them as training points of some nonparametric regression, a plenty of existing non-parametric approaches can be applied to get a reasonable approximation of $\fv_k$ (testing output) at ground truth locations $\phiv_k$ (testing input).  Figure~\ref{reg}(a) shows that if $\phiv_k = 2\tv$ is correctly estimated, then $(\phiv_k,\fv_k)$ should lie on $s_k(t)$. In this case, non-parametric regression can be applied to infer the signal tensity $\fv_k$ at phase position $\phiv_k$ (red dots) using the PiPs (green curve).
Figure~\ref{reg}(b) shows that if $\phiv = 1.5\tv \neq \phiv_k$, $(\phiv,\fv_k)$ deviates from $s_k(t)$. In this case, standard nonparametric methods fail to estimate $\fv_k$ at $\phiv$ (red dots) using PiPs  (green curve) with a large marginal loss.

{
We use the GP regression with squared exponential (SE) kernel to implement this  nonparametric regression step directly on $y_k(\phi_k)$, as it shows robust performance under heavy noise for the stochastic formulation of $y(t)$. We remark here that identical formulation can be derived by performing nonparametric kernel estimator on $f_k(t)$ and treating $ns_k(\bm{t})$ as white noise sampled on a continuous interval.
}

The SE kernel for GP is defined as
\beq
\label{eqn:SE}
\mathbf{k}(t,t') = \beta^{\text{SE}} \exp \left(-\frac{1}{2}   \alpha^{\text{SE}} (t-t')^2  \right),
\eeq
with kernel parameters $\beta^{\text{SE}}$ and ${\alpha}^{\text{SE}}$,
and the posterior of $\yv_k$ given $\phiv_k$, $(\zv_k,\uv_k)$ can be written as
 \begin{align} 
 \ p(\yv_k|\phiv_k,\zv_k,\uv_k)  = \N (\fv_k,\sigma_k^2 I_N )
 = \ \N(K_{NM,k}K_{MM,k}^{-1} \uv_k, \sigma_k^2 I_N),
 \label{eqn:f|u}
\end{align}
according to~\cite{rasmussen2006gaussian}, where the third equality is obtained by computing the marginal distribution of $\yv_k$.
In Eq.~\eqref{eqn:f|u}, $K_{NN,k}$ is a $N\times N$ covariance matrix, $(K_{NN,k})_{ij} = k((\phiv_k)_i,(\phiv_k)_j)$
\footnote{Denote ${(\cdot)}_i$ as the $i$th entry of a vector, and ${(\cdot)}_{ij}$ as the $ij$th entry of a matrix.}
 for $\boldsymbol{\phi_k}\in\R^N$, and similarly $(K_{MM,k})_{ij} =k((\zv_k)_i,(\zv_k)_j)$ for $\zv_k\in\R^{b l}$,  $(K_{MN,k})_{ij} =k((\zv_k)_i,(\phiv_k)_j) = (K_{NM,k})_{ji}$. 

\subsection{Optimization Loss}
\label{sec:analysis}

$p(\yv_k|\uv_k,\phiv_k)$ is higher when $\phiv_k$ are more accurately provided. Based on this fact and independence between different modes, a loss function measuring the differences between the joint probability model $\prod_{k} p(\yv_k|\uv_k,\phiv_k, \zv_k)$ and the ground truth signal $\yv$ is derived to update the latent phase functions.
To be specific, when $a_k(t)\equiv 1$, we have
 \begin{align} 
&p(\yv|\phiv_1,\zv_1,\uv_1,\ldots,\phiv_K,\zv_K,\uv_K)   
=p(\yv|\yv_1,\ldots,\yv_k)  \prod_{k=1}^{K} 
p(\yv_k|\phiv_k,\zv_k,\uv_k) \nonumber   \\
 =& \ \N(\sum_{k=1}^{K} K_{NM,k}K_{MM,k}^{-1} \uv_k,\sum_{k=1}^{K} 
 \sigma_k^2 I_N).\label{eqn:y|f}
\end{align}

We recover the phase function $\phi(t)$ by maximizing the conditional probability of the signal distribution defined in Eq.~\eqref{eqn:y|f}. When $a_k(t)$ is not a constant, Eq.~\eqref{eqn:y|f} can be further modified as following,
\begin{align}
& \mathcal{L}\left( \bm y;\zv_1,\uv_1,\ldots,\zv_K,\uv_K; \phiv_1,\av_1,\ldots,\phiv_K,\av_K
\right) \nonumber  \\
=&  \log p(\bm y | \phiv_1,\zv_1,\uv_1,\ldots,\phiv_K,\zv_K,\uv_K)
= \log \mathcal{N} \left(\bm \mu, \bm \Sigma \right) \label{eq:phi_loss} ,
\end{align}
where
\begin{align}
&  \bm \mu =  \sum_{k=1}^{K} \av_k \odot K_{NM,k}K_{MM,k}^{-1} {\bm u_k} , \nonumber \\
& \bm \Sigma =  \sum_{k=1}^{K}  
\sigma_k^2 I_N . \nonumber
\end{align}
We expect to update $\phi_k(t)$ and amplitudes $a_k(t)$ by maximizing $\mathcal{L}$ given in Eq.~\eqref{eq:phi_loss}.


\subsection{Semi-parametric setting: Parameterization of Phases and Amplitudes}
\label{sec:para}
However, directly updating phase and amplitude functions by maximizing the loss function in Eq.~\eqref{eq:phi_loss}  is highly unstable when no prior information is considered, \eg $\phiv_k(t)$ is smooth and monotone. Thus, we learn the phase function $\phiv_k(t)$ and amplitude function $\av_k(t)$ as sample points from a function, instead of treating $\phiv_k$ and $\av_k$ as independent variable. Among all possible choices, representing $\phi_k(t)$ as low-degree polynomials with order $D$ is effective in practice:
\beq
\label{eqn:para}
\phiv_k = \sum_{d=0}^{D} (B)_{kd}\tv^{d},\quad \text{   and    } \quad
\av_k = \sum_{d=0}^{D_c} (C)_{kd}\tv^{d}.
\eeq
Here $\tv^j$ is the $j$th power of $\tv$.
$B$ and $C$ are $N\times (D+1)$ and $N\times (Dc + 1)$ real matrices, with the $k$th row ($\bv_k$ and $\cv_k$) representing the polynomial coefficients of $\phi_k(t)$ and $a_k(t)$, respectively. Substituting the parameterized forms of $\phiv_k$ and $\av_k$ into Eq.~\eqref{eq:phi_loss}, we can get
\begin{align} \label{eq:para_loss}
 \mathcal{L}\left( \bm y,\zv_1,\uv_1,\ldots,\zv_K,\uv_K; B,C \right) 
= \log \mathcal{N} \left(\bm \mu, \bm \Sigma \right)  ,
\end{align}
where
\begin{align}
& \bm \mu =  \sum_{k=1}^{K} \left( \sum_{d=0}^{D_c}  (C)_{kd} \bm t^{d} \right) \odot K_{NM,k}K_{MM,k}^{-1} {\bm u_k} , \nonumber \\
& \bm \Sigma =  \sum_{k=1}^{K} 
\sigma_k^2 I_N . \nonumber
\end{align}
Note that parameters in $\phiv_k$ are implicitly included in the kernel matrices $K_{NM,k}$ and $K_{NN,k}$. 

The nonparametric setting (Eq.~\eqref{eq:phi_loss}) thus transforms to a semi-parametric setting as Eq.~\eqref{eq:para_loss}.
As a result, the monotonicity and smoothness of $\phiv_k$ is guaranteed, along with a more stabilized performance for this highly non-convex optimization problem. Ignoring the constant terms, we end up with an equivalent MSE loss between $\yv$ and $\bm\mu$
\begin{align}\label{eq:para_loss2}
\mathcal{L}_0\left( \bm y; B,C,\zv_1,\uv_1,\ldots,\zv_K,\uv_K
\right) = ||\yv -  \bm\mu  ||_2 ,
\end{align}
which will be directly optimized by gradient descent methods.

\subsection{Divide-and-Conquer Strategy}
\label{sec:d&c}
In practice, setting $D=2$ provides a reasonable approximation to signals localized in time, because both $\phi_k(t)$ and $a_k(t)$ vary slowly.
For signals that can not be well approximated via lower degree polynomials, a \textit{divide-and-conquer} strategy is applied to obtain a global point estimation for the phase (and amplitude) functions. This divide-and-conquer strategy involves three steps. ($i$) We separate the long signals into multiple localized signal chips. These chips are supported on continuous time intervals which can have intersections, as long as the whole time span of the original signal is fully covered. ($ii$) We update $\phi_k(t)$ (and $\av_k(t)$) for each short chip using polynomial estimation as Eq.~\eqref{eq:para_loss} or Eq.~\eqref{eq:para_loss2}. ($iii$) A robust curve fitting  algorithm~\cite{garcia2010robust,garcia2011fast} is applied to obtain the final global estimation from the previous steps.

If the oscillatory patterns are unknown and need to be updated, to improve the pattern estimate result, Step ($iii$) can be replaced by some more time-consuming variants. This is detailed in the following section.

\section{Estimation of Oscillatory Patterns}\label{sec:Shape}
 In this section, we introduce the way of approximating the oscillating patterns $s_k(t)$. The phase and amplitude functions $\phiv_k$ and $\av_k$ estimated from the previous sections are fixed in this step. We do not aim at closed formulas for ${s}_k(t)$. As introduced in Section~\ref{sec:MI}, it is sufficient to estimate the pattern inducing points $(\zv_k,\uv_k)$ to represent the non-oscillating pattern $\hat{s}_k(t)|_{[a,b]}$. 
When amplitude and phase functions are given, shape function estimation has been studied thoroughly in previous works, like \cite{FMMD,xu2018recursive,1DSSWPT,MMD}. There are no quantitative criteria to measure how well the shape function estimate performs when the amplitude and phase function estimate is not very good. These methods achieve good performance when the inferred amplitude and phase functions are close to the ground truth. 
%

\subsection{Summarize Output from Localized Chips}\label{sec:two-scale}
In this section, we illustrate two variants in Step ($iii$) of  Section~\ref{sec:d&c} that can improve the accuracy of pattern estimation.

First, since different signal chips generate instantaneous information estimates with distinct qualities and these qualities of instantaneous information can be partially manifested by the final loss of Eq.~\eqref{eq:para_loss} or Eq.~\eqref{eq:para_loss2}, we propose a loss-selective variant of Step ($iii$).  To be specific, instead of feeding all the chips' output into a smooth curve fitting module to generate the pattern estimate, we select chips with relatively low loss value for the pattern estimation. For this end, two constants are prefixed: a threshold value $\tau_1$ to admit all chips that satisfies $\mathcal{L} < \tau_1$; if all chips has loss larger than $\tau_1$, then we use another quantile constant $\tau_2$ to select a portion of chips with the lower loss value. As a result, we can enhance the possibility that better estimated instantaneous information is applied, while instantaneous information with lower qualities is discarded,  to update the oscillatory patterns.

However, empirical experiments show that lower loss value from Eq.~\eqref{eq:para_loss} or Eq.~\eqref{eq:para_loss2} does not always guarantee a better pattern recovery. 
We further propose a more expensive loss-selective variant of Step ($iii$) by computing and summarizing the distribution of the final losses generated by the chips in the next iteration step for each chip in the current iteration step.
In practice, we set the average of final losses in the next iteration as our fresh chip quality indicator and then apply the identical chip selective scheme work as illustrated in the aforementioned paragraph.
Although it's not always the case that this fresh indicator guarantees a better guideline for pattern updating, this variant can guarantee a lower instantaneous information loss in the next round of iteration.
We note that this variant involves extensive computing of the trial-and-errors for selecting ideal chips, thus is highly time-consuming compared to the first variant.

\section{Overview of PiPs}\label{sec:pgpr}
In this section, an overview of the whole algorithm is presented.
PiPs repeatedly applies alternatively updates between spectral information and oscillatory patterns until convergence. The overall loss function of PiPs can be written as:
\begin{align}\label{eq:final_loss}
& \mathcal{L}\left( \bm y;  \bm B, \bm C,\bm z_1,{s}_1(\cdot),\ldots,\bm z_K,{s}_K(\cdot) \right)  = \log \mathcal{N} \left(\bm \mu, \bm \Sigma \right) , \\
\text{where  }\ \ & \bm \mu =  \sum_{k=1}^{K} \left( \sum_{d=0}^{D_c}  (C)_{dk} \bm t^d \right) \odot K_{NM,k}K_{MM,k}^{-1} {s_k(\bm z_k)} , \nonumber 
\text{and  }\ \bm \Sigma = \sum_{k=1}^{K} \sigma_k^2 I.
\end{align}
The goal is to maximize the conditional probability of $p(\bm y | \bm z_1,\ldots,\bm z_K)$ with respect to the parameters in phase, amplitude and shape functions, i.e.,
\begin{equation}\label{eq:final_objective}
\operatornamewithlimits{\text{sup}}\limits_{s_k(\cdot) \in \mathcal{S}} \max_{\bm B, \bm C}\mathcal{L} \left( \bm y;  \bm B, \bm C,\bm z_1,{s}_1(\cdot),\ldots,\bm z_K,{s}_K(\cdot) \right) 
,
\end{equation}
where $\mathcal{S}$ indicates the set of predefined non-oscillatory patterns supported on $[0,1]$. $\bm B$ and $\bm C$ are the matrix form of phase and amplitude functions in Eq.~\eqref{eqn:para}. The pseudo-code of the proposed algorithm is given in Algorithm~\ref{alg:pgp}. We only put the most basic update procedure in the pseudo-code. We implement the gradient descent using Adam~\cite{kingma2014adam} with learning rate set to 0.005. 

\begin{figure}[t]
    \begin{center}
    \begin{tabular}{cc}
      \includegraphics[height=2.8in]{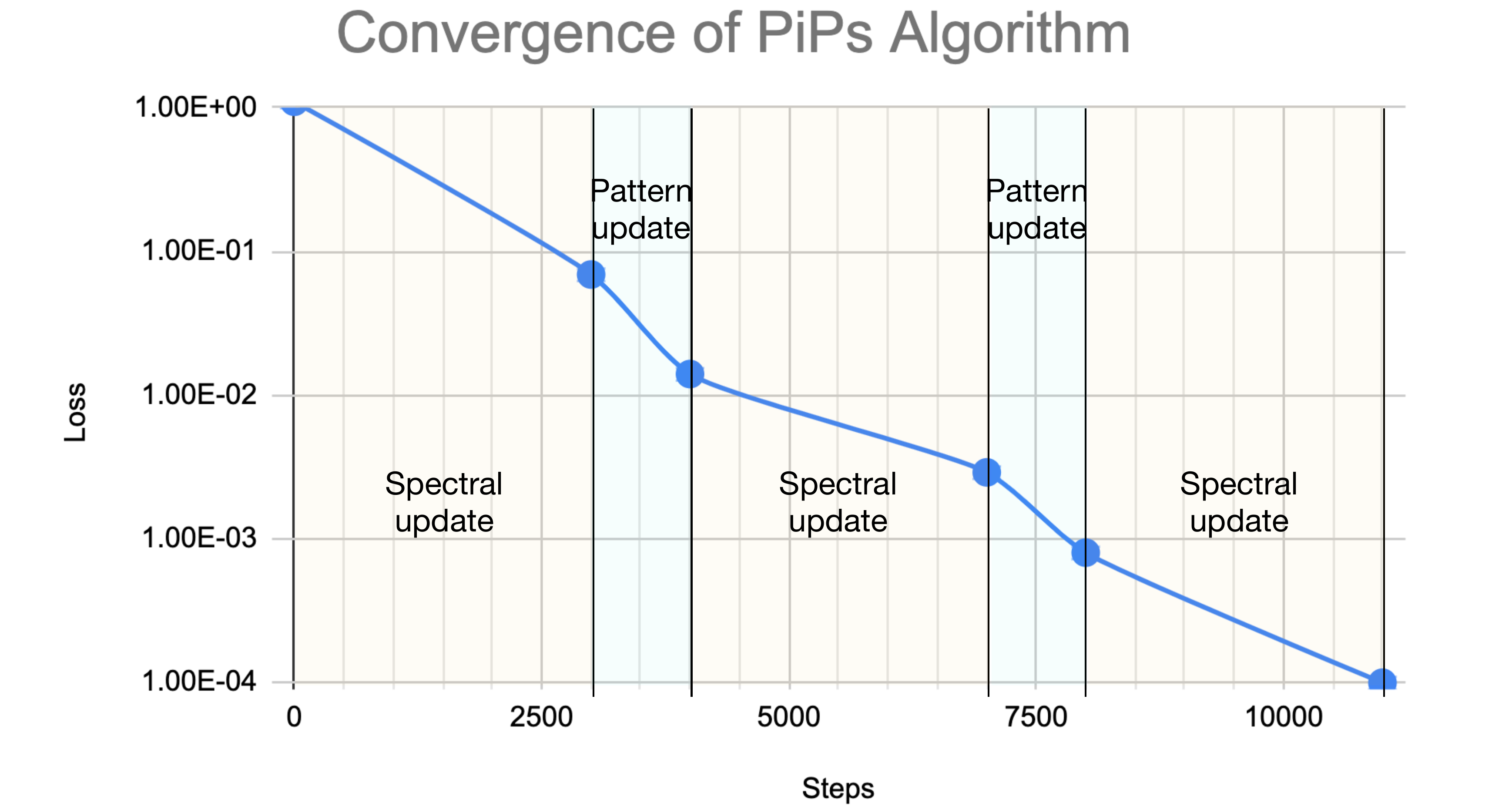} 
    \end{tabular}
\end{center}
\caption{Illustration for the convergence process of Algorithm~\ref{alg:pgp} for the PPG sparse dataset.}
\vspace{-0.3cm}
\label{conv}
\end{figure}

{
For each outer loop, we first update the parameterized phase and amplitude functions for each component using the GP based gradient descent method as introduced in Section~\ref{sec:MI}. Secondly, the oscillation patterns are updated with RBDR as illustrated in Section~\ref{sec:Shape}.
For sparsely sampled signals (less than 1000 sample points as with the sparse PPG example), PiPs usually converges with two or three outer loop iterations with phase updating part converges within 3000 steps. An illustrative convergence pattern is shown in Figure~\ref{conv}, whereas we can observe the pattern updating stage is effective and crucial to the overall optimization process. Theoretical analysis of this alternative approach will be treated as a future work of this approach.
}

\begin{algorithm}[]
{\bf Input}:  $N$ measurements of $(\tv,\yv)$, 
the number of components $K$, 
the polynomial degrees $D_c$ and $D$,
input of PIPs $\zv_k$ (k=1,\ldots,K) and the accuracy parameter $\epsilon$.

{\bf Initialization}: Initialize the estimates of oscillatory patterns $s_k(\cdot)$, the phase and amplitude parametrized matrices $B$ and $C$, and set $iter$ = 0. For long signals, divide the long signal into short chips (Section~\ref{sec:Shape}).

\While{$iter < MaxIter$ and model not converge}{

Compute $\bm u_k = s(\bm z_k)$ ($k=1,\cdots,K$) for each component.

\For {each chip}{
\For {$i=1$ to $Iter_{\phi, a}$}{



Fix $s_k(\cdot)$, compute $\Delta \bm b_k = \frac{\nabla \mathcal{L}}{\nabla \bm a_k} \frac{\nabla \bm a_k}{\nabla \bm b_k}$, $\Delta \bm c_k = \frac{\nabla \mathcal{L}}{\nabla \bm \phi_k} \frac{\nabla \bm \phi_k}{\nabla \bm c_k}$ for $k=1,\cdots, K$. (Section~\ref{sec:para}).

Update $\bm b_k = \bm b_k + \gamma \Delta \bm b_k$, and $\bm c_k = \bm c_k + \gamma \Delta \bm c_k$.
}

Compute $\bm a_k = \sum_{d=0}^D B_{dk} \bm t^d$ and $\bm \phi_k = \sum_{d=0}^D C_{dk} \bm t^d$.
}

Update oscillatory patterns $s_k(\cdot)$ for $k=1,\cdots,K$ with selected chips. (Section~\ref{sec:Shape}).}

{\bf Output}: estimates of the pattern inducing variables $\uv_k$ representing $s_k(t)$, 
and the latent variables $\phiv_k$ and $\av_k$ representing $\phi_k(t)$ and $a_k(t)$, respectively.

\caption{(PiPs) Note that an alternative is to initialize the phase $\{\phiv_k\}$ and amplitude $\{ \av_k\}$ as fixed, while updating the oscillatory patterns first. The order of updating which component first depends on the signal and prior knowledge. Note that if the initialization of oscillatory patterns is better than those of amplitude and phase functions, we update oscillatory patterns first. Since the problem is not convex, the global convergence analysis of the proposed algorithm would be interesting future work.} \label{alg:pgp}
\end{algorithm}

In many applications, e.g. ECG and PPG data analysis, heuristic properties of the physical system are often available and we know the rough range of instantaneous frequencies $\omega_k(t)$. Hence, we can apply a band-pass filter to $f(t)=\sum_{k=1}^{K} a_k(t)s_k(\phi_k(t))=\sum_{k=1}^K \sum_n \widehat{s}_k(n) a_k(t) e^{\imath n\phi_k(t)} $ with Fourier expansion on the shape function. Then, we estimate amplitude and phase of $\widehat{s}_k(1) a_k(t) e^{\imath\phi_k(t)}$ in a certain frequency band using traditional time-frequency analysis methods \cite{Daubechies1996,YANG2017}. Generally, $\widehat{s}_k(1)$ is much larger than the rest Fourier coefficients. Thus, by setting $K$ to $1$, Synchrosqueezed-based methods~\cite{daubechies2011synchrosqueezed} can be directly applied without a band filter. Another initialization method is to directly set $\uv_k=\sin(\zv_k)$. Since we adopt local patch segmentation in Section \ref{sec:MI}, components decomposed by Fourier series expansion become approximately orthogonal to each other in a short time period ($\{\widehat{s}_k(1) a_k(t) e^{\imath\phi_k(t)}\}_{k=1}^K$). Hence, Algorithm~\ref{alg:pgp} can recover the amplitude and phase functions corresponding to $\{\widehat{s}_k(1) \}_{k=1}^K$ since they usually have the largest $K$ magnitude.

%
%

\section{Experiments}
\label{sec:num}
\vspace{-5pt}

In this section, we provide numerical examples to demonstrate the performance of PiPs\footnote{Code is available on \url{https://github.com/JierenXu/PiPs}.}, especially in the case of super-resolution and adaptive time-frequency analysis. Optimization problems in all examples are solved by Adam~\cite{goodfellow2016deep} aiming at better local minimizers. We choose degree-$1$ (or degree-$2$ when specified) polynomials to approximate local amplitude and phase functions in these optimization problems. The hyperparameters of PiPs are set as follows: noise level $\sigma = 10^{-0.8}$, $\alpha^{\text{SE}}=2\times 10^{3} $, and $\beta^{\text{SE}} = 1$. In the local patch analysis, we generate signal patches such that each patch contains approximately $3$ to $10$ periods. In the tests for super-resolution, we repeat the same test with $10$ noise realizations to use the expectation and variance of estimation error to measure the performance of different algorithms.
$\Delta \omega$ and $\Delta \phi$ denote the point-wise estimation error. We also created a set of oscillation patterns with non-trigonometic shapes to facilitate testing. Whenever appeared, we refer
\begin{equation*}
    s_1^{eg}(t) = 
    \begin{cases}
     \frac{16}{\pi}t - \frac{4}{\pi}, \text{ when } 0 \le t < 0.5,\\
        \frac{12}{\pi}- \frac{16}{\pi}t, \text{ when }  0.5 \le t < 1, 
    \end{cases}
    \end{equation*}
and
    \begin{equation*}
    s_2^{eg}(t) = 
    \begin{cases}
    -8t +3, \text{ when } 0.25 \le t < 0.5, \\
     8t-7, \text{ when }  0.75 \le t < 1,\\
       1, \text{  otherwise.} 
    \end{cases}
\end{equation*}
These shape functions have been visualized in Fig. \ref{fig5} (b) and (c).

\subsection{Super-resolution spectral estimation}
\label{exp:const}

There has been substantial research for the super-resolution problem that aims at estimating time-invariant amplitudes and frequencies in a signal $f(t)=\sum_{k=1}^K a_k e^{i\omega_k t}$ with $a_k>0$, $\omega_k>0$, and $\{\omega_k\}$ are very close. Among many possible choices, the baseline models might be MUSIC~\cite{schmidt1986multiple}, ME~\cite{burg1972relationship,georgiou2001spectral,georgiou2002spectral}, and ESPRIT~\cite{roy1989esprit}. Hence, we will compare PiPs with these methods\footnote{Code from \url{http://people.ece.umn.edu/~georgiou/files/HRTSA/SpecAn.html}.} to show the advantages of PiPs. Although the Fourier transform usually fails~\cite{schmidt1986multiple} to identify $\{a_k\}$ and $\{\omega_k\}$, we use its results as the initialization for PiPs. 

\vspace{-10pt}
\begin{figure}[t]
  \begin{center}
    \begin{tabular}{cccc}
      \includegraphics[height=1.8in]{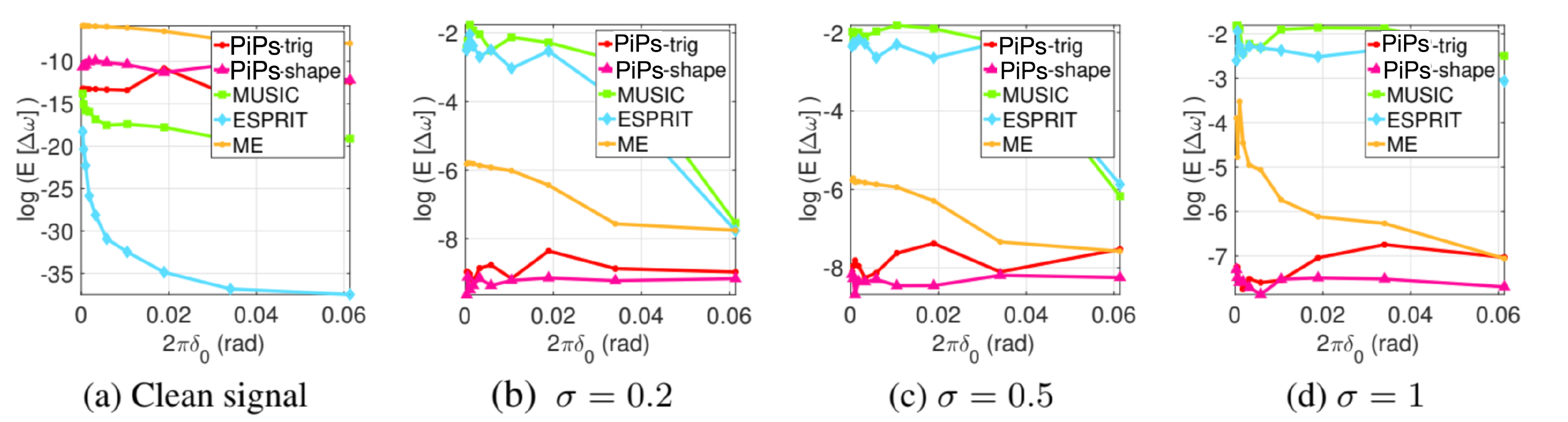} 
    \end{tabular}
  \end{center}
  \caption{ \footnotesize{Frequency estimate (absolute) error of $f^{\{1\}}(t) = \cos(2\pi\omega_1 t) + \sin(2\pi \omega_2 t)$, where $\omega_1 = \nicefrac{38.8}{1024}$ and $\omega_2 = \nicefrac{(38.8+\delta_0)}{1024}$ with different $\delta_0$ and white noise $\mathcal{N}(0,\sigma^2)$.}}
  \label{fig1}
\end{figure}

  \begin{figure}[t]
    \begin{center}
    \begin{tabular}{cc}
      \includegraphics[height=4in]{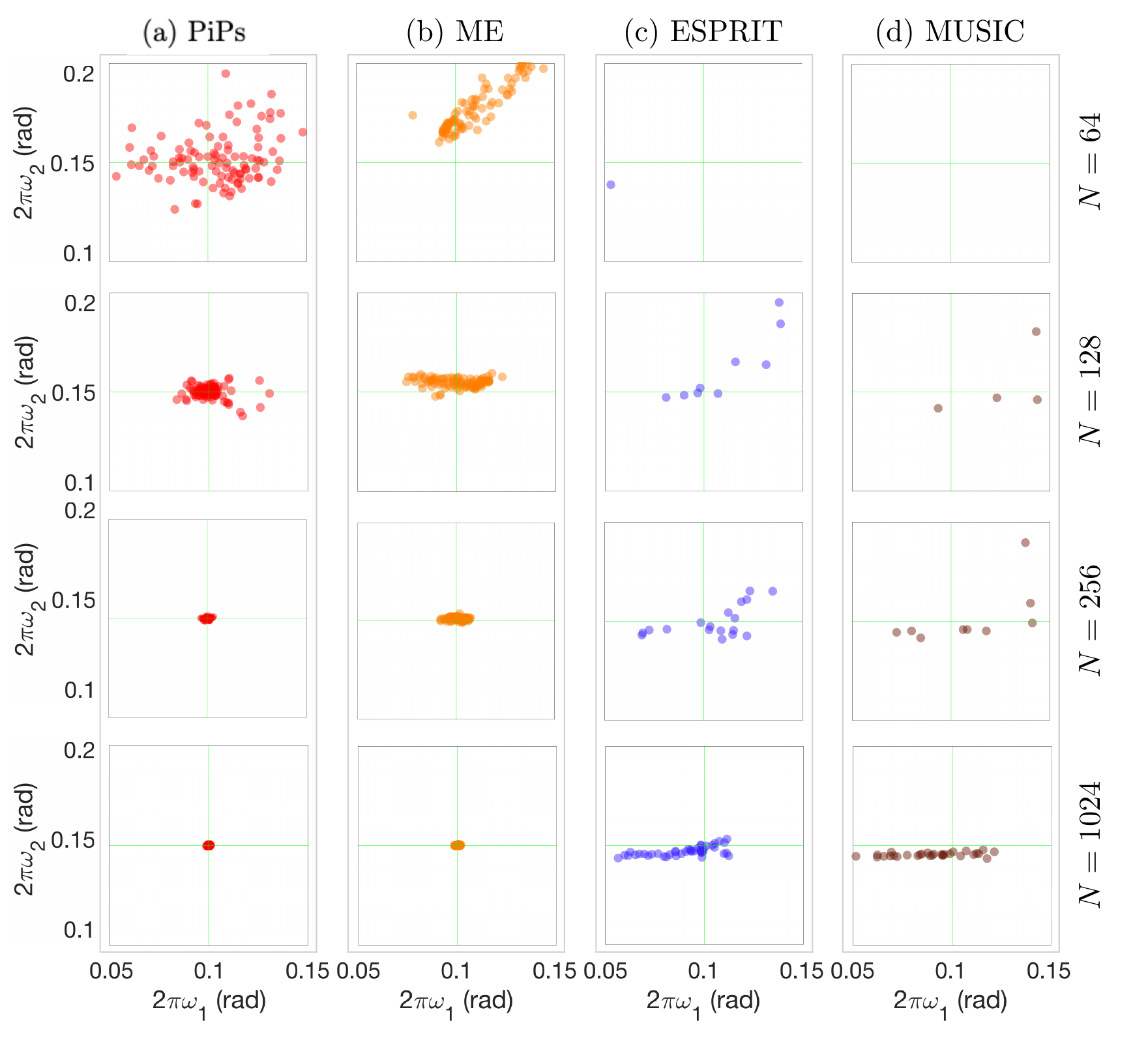} 
      \includegraphics[height=3.1in]{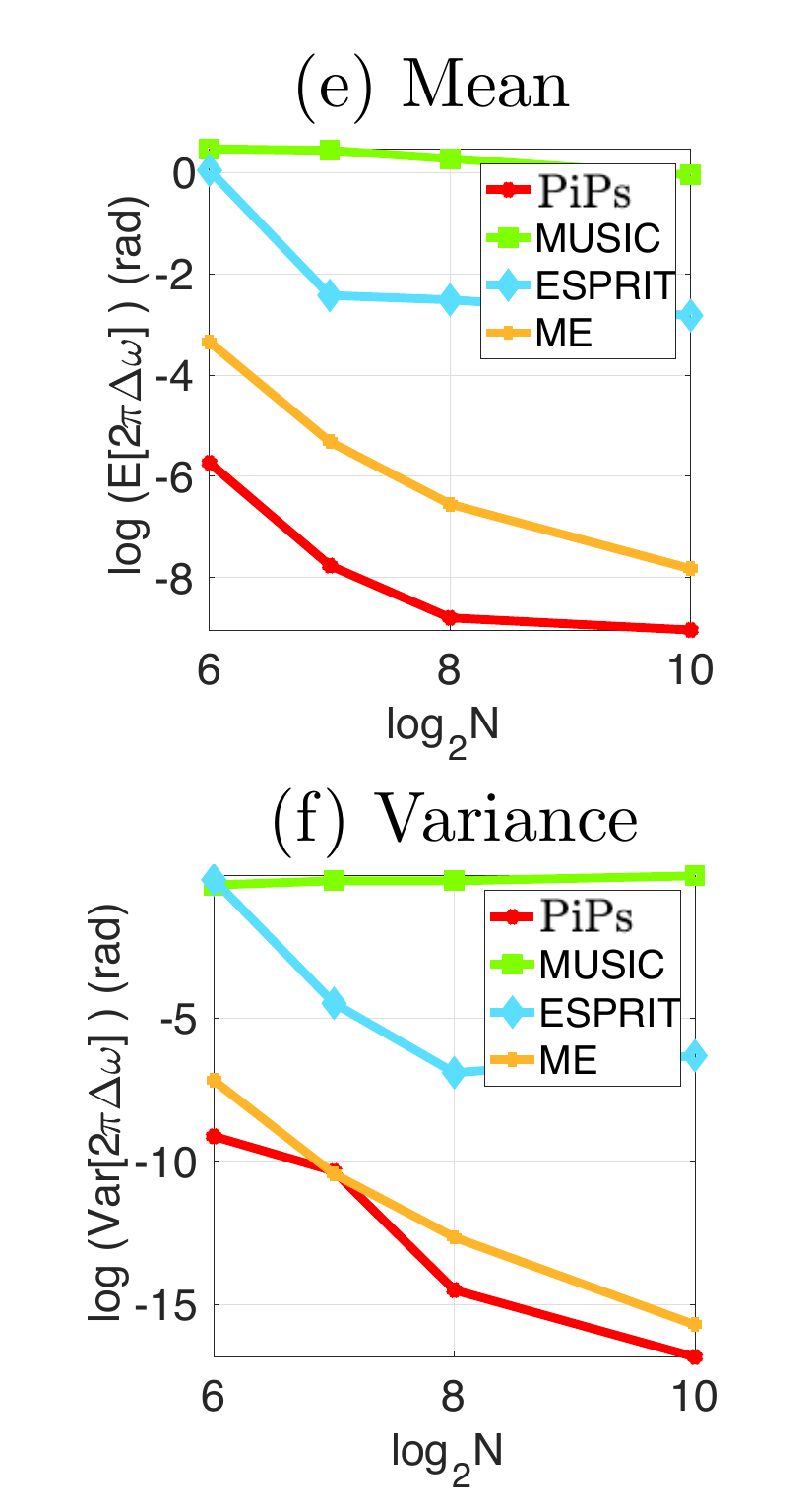}
    \end{tabular}
\end{center}
\vspace{-2pt}
  \caption{ \footnotesize{Left: results of $f^{\{2\}}$ with different number of samples $N=64$, $128$, $256$, and $1024$ from top to bottom, and by different methods in an order of PiPs, ME, ESPRIT, and MUSIC from left to right. The ground truth frequencies are $({2\pi\omega}_1, {2\pi\omega}_2)=(0.1,0.15)$. $100$ tests with different noise realization were performed and  the estimated frequencies are visualized in a $2$D domain centered at the ground truth.  Right: the expectation and variance of estimation errors for different methods and numbers of samples.}}
  \label{fig2}
\end{figure}

\paragraph{Accuracy and robustness with different spectral gaps} In this experiment, we use $f^{\{1\}}(t) = s_1(2\pi\omega_1 t) + s_2(2\pi \omega_2 t)+\mathcal{N}(0,\sigma^2)$, where the two instantiations of $s_1(\cdot)$ and $s_2(\cdot)$ are chosen as follows: 
\begin{itemize}
    \item 
$(i)$ $s_1(t) = \cos(t) $ and $s_2(t) = \sin(t)$ for standard super resolution comparison to other baseline methods. The estimation results by PiPs are denoted as PiPs-trig and are visualized  as red lines in Fig.~\ref{fig1}.
\item $(ii)$ $s_1(t) = s_1^{eg}(t)$ and $s_2(t) = s_2^{eg}(t)$ for super resolution comparison with special oscillation patterns. The estimation results by PiPs are denoted as PiPs-shape and are visualized  as pink lines in Fig.~\ref{fig1}.
\end{itemize}
In these two examples, $\omega_1 = \nicefrac{38.8}{1024}$ and $\omega_2 = \nicefrac{(38.8+\delta_0)}{1024}$; $\delta_0$ varies from $\nicefrac{0.05}{1024}$ to $\nicefrac{10}{1024}$; and the noise variance is $\sigma^2=10^{-1.6}$. The sampling rate is $1$ Hz and the number of samples is $N=100$ in this example. Fig.~\ref{fig1} shows the frequency estimation accuracy of PiPs, MUSIC, ESPRIT, and ME. 

{
As we can see, PiPs achieves machine accuracy in the noiseless case and is much more accurate than other methods in all noisy cases.  It also reaches almost the same accuracy for the trigonometric (red) and shaped (pink) instantiations in $(i)$ and $(ii)$.
All baseline methods can not directly apply to instantiation $(ii)$ with non-trigonometic oscillation patterns besides PiPs.}

\paragraph{Accuracy and robustness with different sampling rates} In this experiment, we set $f^{\{2\}}(t) =  \sum_{k=1}^2 a_k \sin( \omega_k t)$ with $a_1 = 0.5, a_2 =1 $, $\omega_1 = 0.1$, and $\omega_2 = 0.15$. The sampling rate of this signal is still $1$Hz and the numbers of samples are $N=64$, $128$, $256$, and $1024$ to generate four sets of test data. There are two different kinds of noise to generate noisy test data: 1) white Gaussian noise $\mathcal{N}(0,0.35)$ is directly added to $f(t)$; 2) a stochastic process in $t$ with i.i.d. uniform distribution in $[0,2\pi]$ is added to phase functions $\{\omega_kt\}_{1\leq k\leq 2}$. Fig.~\ref{fig2} summarizes the results of frequency estimates in this experiment. ESPRIT and MUSIC lose accuracy in all tests. PiPs and ME achieve high accuracy when the number of samples is large and PiPs is slightly better than ME in terms of accuracy and estimation bias.

\subsection{Estimation of time-variant frequencies}
\label{exp:variant}
 
 In this section, we show the capacity of PiPs for estimating close and crossover time-varying instantaneous frequencies. An adaptive time-frequency analysis algorithm, ConceFT~\cite{Daubechies20150193}), is used as a comparison.
 And local approximation degree is set to $d=2$ in this section.
 
\begin{figure}[t]
  \begin{center}
       \includegraphics[height=2.3in]{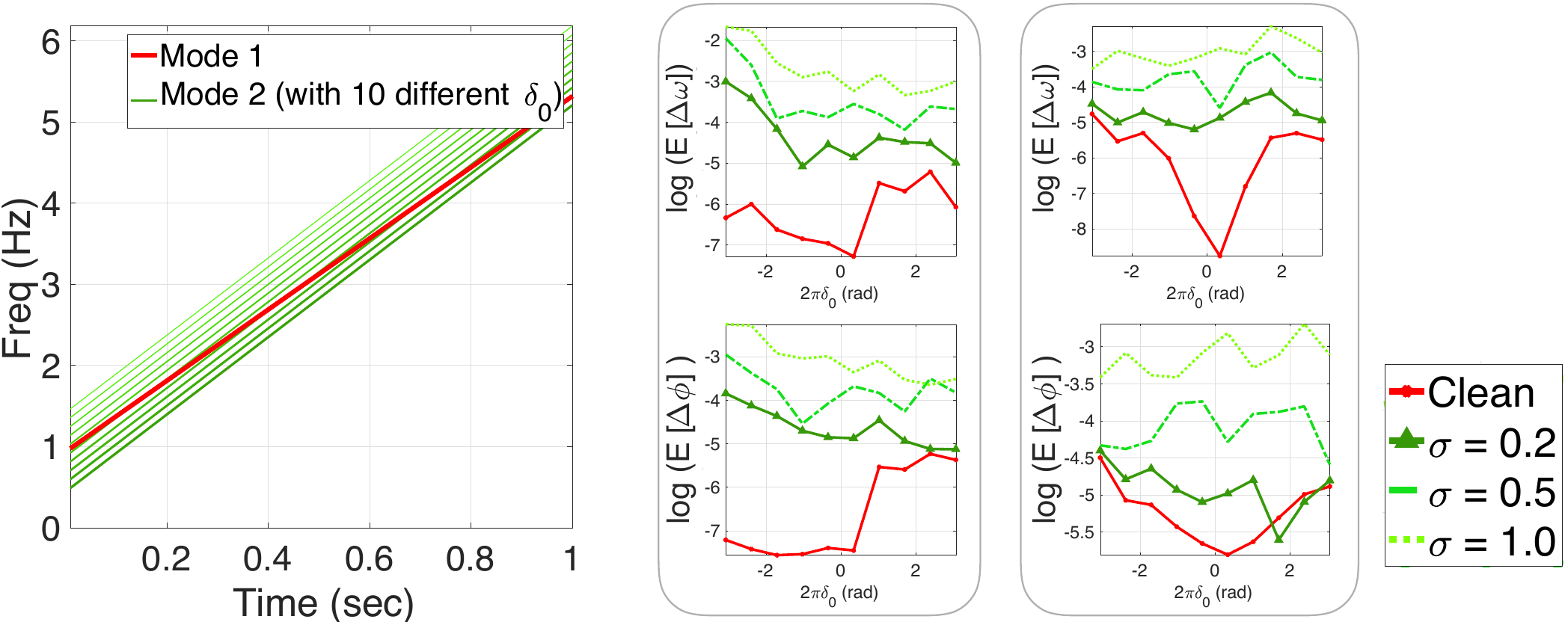}  \\
  \small{  \hspace{-45pt}   (a) $\omega_1(t)$ and $\omega_2(t)$ in time-frequency domain \hspace{5pt} (b) Error for $s_k = \sin$ \hspace{5pt}  (c) Error for $s_k = s^{eg}_k$}
  \end{center}
  \vspace{-8pt}
  \caption{\footnotesize{Short signal $f^{\{3\}} (N=100)$ with linear frequencies. 
  (a)  visualizes all the instantaneous frequencies of our synthetic components as the spectral gap parameter $\delta_0$ takes the values $(i-5)/10.24$ for $i=0,1,...,9$.
  (b) is the estimation error for frequency (top) and phase (bottom) estimates when $s_k = \sin$; (c) is for $s_k = s^{eg}_k$.}}
  \label{fig3}
  \vspace{-10pt}
\end{figure}

\vspace{-10pt}
\paragraph{Close frequencies and phase estimation error} We use
$f^{\{3\}}(t) = s_1(2\pi(\nicefrac{10}{10.24} t+\nicefrac{230}{10.24^2} t^2)) + s_2(2\pi ((\nicefrac{10}{10.24} +\delta_0)t+\nicefrac{250}{10.24^2} t^2))$, where the two instantiations of $s_1(\cdot)$/$s_2(\cdot)$ are $(i)$ $s_1 = \cos $/$s_2 = \sin$ (Fig.~\ref{fig3}(b)) and $(ii)$ $s_1 = s_1^{eg}$/$s_2 = s_2^{eg}$ as in Fig.~\ref{fig3}(c). 
$\delta_0$ varies from $-\nicefrac{5}{10.24}$ to $\nicefrac{5}{10.24}$. The white noise $\sigma_0$ has standard deviation $\{0,0.2,0.5,1\}$. 
We apply short-time Fourier transform~\cite{griffin1984signal} to identify rough estimates of instantaneous frequencies and use them as the initialization in this test. When instantaneous frequencies are very close, the initialization is very poor; however, PiPs still can identify instantaneous frequencies and phases with reasonably good accuracy.
The result is summarized in Fig.~\ref{fig3}.

Fig.~\ref{fig3}(a) is the ground truth time-frequency representation of ten tested signals with different value of $\delta_0$ on $\omega_2(t)$. The difference between $\omega_2(t)$ (green line) and $\omega_1(t)$ (red line) are pretty difficult to be detected by existing time-frequency methods.
The log error of the point-wise averaged frequency estimate is shown in the first row of Fig.~\ref{fig3} (b) and (c) on different noise levels $\sigma_0$.
The log error of point-wise averaged phase estimate (bottom row) is consistently small as $\delta_0$ changes. Under a large noise case with $\sigma_0 = 1$, PiPs controls the phase error approximate or below the level of $0.05$. 
Existing time-frequency analysis methods usually estimate instantaneous frequencies first and then integrate them to obtain instantaneous phases, which suffers from accumulated error. However, PiPs has no accumulated error.

\begin{figure}[t]
  \begin{center}
   \includegraphics[height=2.3in]{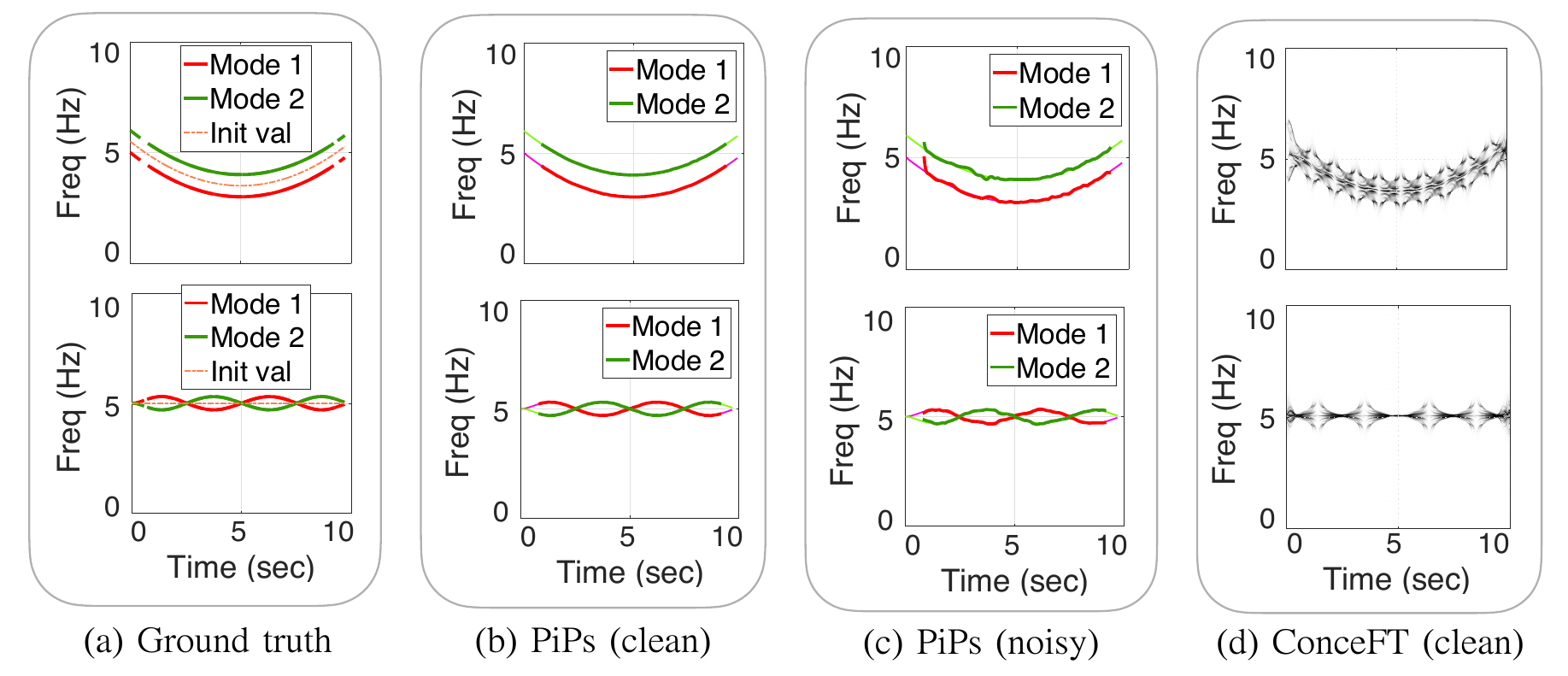}  \\
  \end{center}
  \vspace{-7pt}
  \caption{\footnotesize{Instantaneous frequency estimates for signals with close and crossover frequencies. (a) ground truth instantaneous frequencies and initialization of PiPs. (b) estimated instantaneous frequencies for clean signals. (c) estimated instantaneous frequencies for noisy signals. (d) time-frequency distribution by ConceFT. }}
  \label{fig4}
  \vspace{-10pt}
\end{figure}

\vspace{-5pt}
\paragraph{Close and crossover frequencies } In this experiment, we generate a signal consisting of two components with close instantaneous frequencies and a signal with two crossover instantaneous frequencies. Fig.~\ref{fig4} visualizes the ground truth instantaneous frequencies, the time-frequency distribution by ConceFT, the initialization, and the estimation results of PiPs. ConceFT cannot visualize the instantaneous frequencies even if in the noiseless case. We average out the energy distribution of ConceFT to obtain the initialization of PiPs. 
Although the initialization is very poor, PiPs is still able to estimate the instantaneous frequencies with a reasonably good accuracy no matter in clean or noisy cases.
%
When the number of components $K$ is known, we generally can average out the energy band to obtain one instantaneous frequency function and initialize all instantaneous frequencies in PiPs using this function from empirical observations. Similar initialization strategy is used in the following examples.

\vspace{-2pt}

\begin{figure}[t]
  \begin{center}
   \includegraphics[height=2.5in]{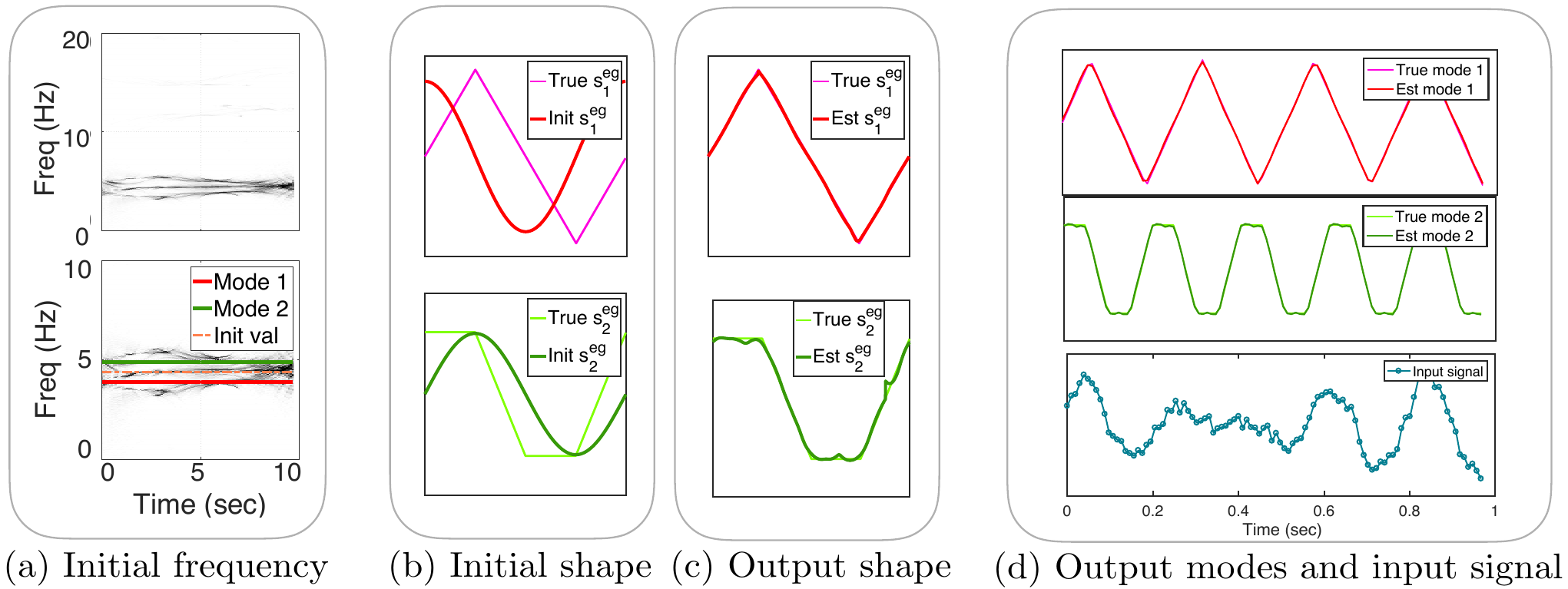}  \\
  \end{center}
  \vspace{-7pt}
  \caption{\footnotesize{PiPs is applied to estimate the amplitude, phase, and shapes of a synthetic signal $f^{\{6\}}(t)$ consisting of two components. (a) the time-frequency distribution of $f^{\{6\}}(t)$ by ConceFT in two different frequency ranges. ConceFT cannot reveal the ground truth instantaneous frequencies (in red and green). But we can initialize PiPs by averaging out the distribution (see the dash pink line). (b) and (c) the ground truth shape functions and their estimates. (d) the noisy signal $f^{\{6\}}(t)$ and the reconstructed components by PiPs.
 }}
  \label{fig5}
\end{figure}
\begin{figure}[t]
  \begin{center}
   \includegraphics[height=4.5in]{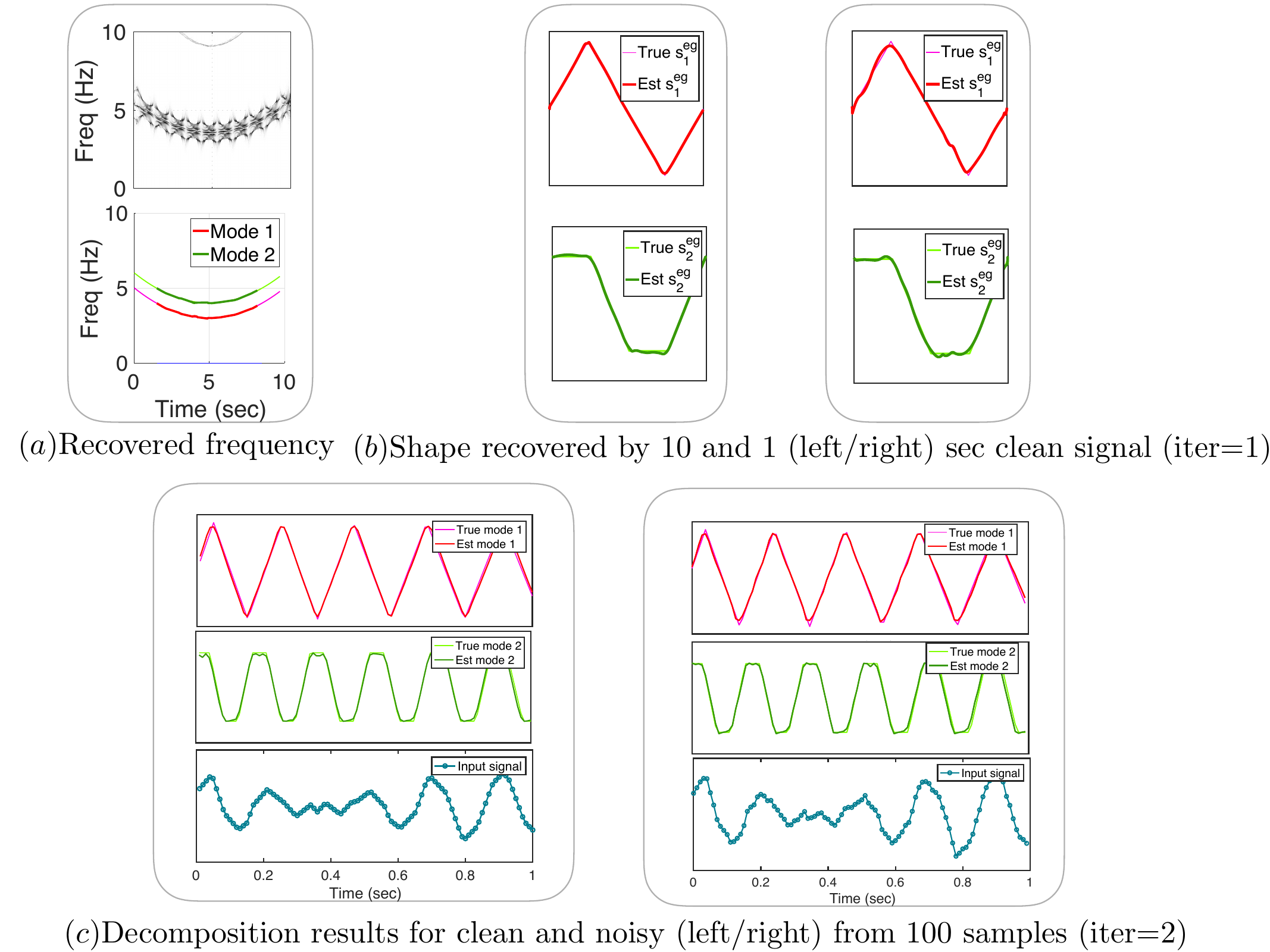} \\
  \end{center}
  \vspace{-2pt}
  \caption{\footnotesize{PiPs is applied to estimate the amplitude, phase, and shapes of a synthetic signal $f^{\{7\}}(t)$ consisting of two components. (a) the time-frequency distribution of $f^{\{7\}}(t)$ by ConceFT in two different frequency ranges. ConceFT cannot reveal the ground truth instantaneous frequencies (in red and green). But we can initialize PiPs by averaging out the distribution (see the dash pink line). (b)the ground truth shape functions and their estimates. In this case we apply the divide-and-conquer trick due to the non-linearity of frequency functions and take 10 signal chips whereas each for 1 second. We observed that when more chips are applied, the shape are more accurately estimated.  (c) the noisy signal $f^{\{7\}}(t)$ and the reconstructed components by PiPs.
 }}
  \label{fig9}
  \vspace{-15pt}
\end{figure}

\begin{figure}[t]
  \begin{center}
   \includegraphics[height=3.2in]{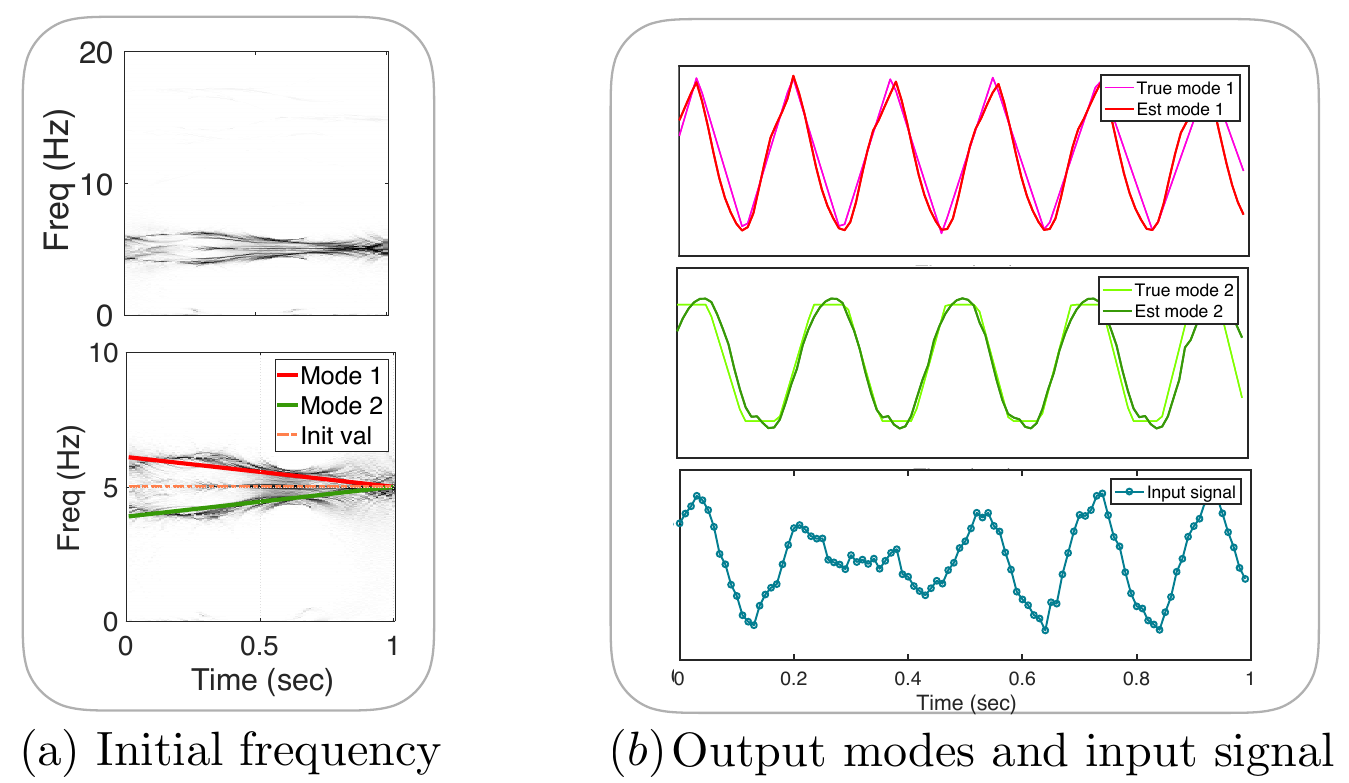}  \\
  \end{center}
  \vspace{-2pt}
  \caption{\footnotesize{PiPs is applied to estimate the amplitude, phase, and shapes of a synthetic signal $f^{\{8\}}(t)$ consisting of two components. (a) the time-frequency distribution of $f^{\{8\}}(t)$ by ConceFT in two different frequency ranges. ConceFT cannot reveal the ground truth instantaneous frequencies (in red and green). But we can initialize PiPs by averaging out the distribution (see the dash pink line). (b) the noisy signal $f^{\{8\}}(t)$ and the reconstructed components by PiPs.
 }}
  \label{fig10}
  \vspace{-15pt}
\end{figure}

\subsection{Estimation of amplitudes, phases, and shapes simultaneously}

\label{exp:shape}

Finally, we apply PiPs to estimate amplitudes, phases, and shapes simultaneously from a single record. First, we generate a synthetic example $f^{\{6\}}(t) = \sum_{k=1}^{2} s_k^{eg}(\omega_k t) +\mathcal{N}(0,0.2)$, where $\omega_1 = \nicefrac{3.88}{1.024}$, $\omega_2 = \nicefrac{4.88}{1.024}$, and the shapes are visualized in Fig. \ref{fig5}. The sampling rate for this signal is $100$ Hz and we sample it at $100$ locations. The shape estimates are initialized as $\cos$ and $\sin$ for the first and second components, respectively. The frequency estimates are initialized as {\em one} constant centered in the peak spectrogram by ConceFT (see Fig. \ref{fig5} (a)). As we can see in Fig. \ref{fig5} (b) and (c), PiPs is able to estimate shape functions with a reasonably good accuracy and the reconstructed components match the ground truth components very well. 

In Fig. \ref{fig9} and \ref{fig10}, a similar initialization strategy is applied to two more examples, $f^{\{7\}}(t)=\sum_{k=1}^{2} s_k^{eg}(\omega_k^{smooth}(t) ) +\mathcal{N}(0,0.2)$ and $f^{\{8\}}(t)=  \sum_{k=1}^{2} s_k^{eg}(\omega_k^{near}(t) ) +\mathcal{N}(0,0.2)$, respectively. Here
\begin{align}
& \omega_1^{smooth}(t)= 0.08t(t-10), \text{  }\omega_2^{smooth}(t)= 0.08t(t-10) + 1,\nonumber \\
& \omega_1^{near}(t)  = t + 4, \text{  and } \omega_2^{near}(t)=  - t + 6.\nonumber
 \end{align}
In other words, signal $f^{\{7\}}(t)$ and $f^{\{8\}}(t)$ both have more difficult frequency time-frequency representation to resolve, on with non-linearity and one with contact frequency curves. The results in Fig. \ref{fig9} and \ref{fig10} show that PiPs can still obtain sharp mode decomposition results for both clean and noisy cases with sparsely sampled data points in these challenging examples.

In the last example, we apply PiPs to a real signal from photoplethysmogram (PPG) (see Fig. \ref{fig6}). 
The shape estimates are still initialized as $\cos$ and $\sin$ for the two components, and $N=100$ samples are involved. The PPG signal contains two components corresponding to the health condition of the heart and lungs in the human body, where Fig. \ref{fig6}(b) shows the mode decomposition result. As can be seen, two modes with highly domain-specific patterns are accurately recovered under the naive cosine and sine oscillation pattern initialization from these 100 samples.

\section{Conclusion}
{
This paper proposed a novel alternatively learning scheme (PiPs) between spectral information and periodical patterns to address several oscillatory data analysis problems, including signal decomposition, super-resolution, and signal sub-sampling. The method achieves state-of-the-art results for noisy and sparsely sampled cases on several datasets, and demonstrates its potentials in real world applications. Though numerical convergence of the proposed method has been observed, an interesting future direction is to analyze the convergence theoretically, especially statistical analysis in the presence of noise.
}

\vspace{.2cm}
{\bf Acknowledgments.} H. Y. was partially supported by the NSF Award DMS-2244988 and DMS-2206333, and the Office of Naval Research Award N00014-23-1-2007.
 
\bibliographystyle{unsrt}
\bibliography{subtex/references} 


\end{document}